%% file: coling2020.tex
\documentclass[11pt]{article}
\usepackage{coling2020}
\usepackage{times}
\usepackage{url}
\usepackage{latexsym}
\usepackage{xcolor}
\usepackage{ dsfont }
\usepackage{graphicx}
\usepackage{amsmath}
\usepackage{graphicx}
\usepackage{pdfpages}

\usepackage[T1]{fontenc}
\usepackage[font=small,labelfont=bf,tableposition=top]{caption}

\DeclareCaptionLabelFormat{andtable}{#1~#2  \&  \tablename~\thetable}
\usepackage{amssymb}
\usepackage{subfig}

\title{	Finding the Evidence: Localization-aware Answer Prediction for Text Visual Question Answering}

\author{Wei Han\thanks{* These authors contributed equally to this work}\\
  MediaTek, Singapore \\
  {\small \texttt{Wei.Han@mediatek.com} } \\\And
  Hantao Huang\footnotemark[1] \\
  MediaTek, Singapore \\
  {\small \texttt{Hantao.Huang@mediatek.com}} \\\And
  Tao Han \\
  MediaTek, Singapore \\
  {\small \texttt{TaoH.Han@mediatek.com}} \\}

\date{}

\begin{document}
\maketitle

\begin{abstract}
Image text carries essential information to understand the scene and perform reasoning. Text-based visual question answering (text VQA) task focuses on visual questions that require reading text in images.
Existing text VQA systems generate an answer by selecting from optical character recognition (OCR) texts or a fixed vocabulary. Positional information of text is underused and there is a lack of evidence for the generated answer. As such, this paper proposes a localization-aware answer prediction network (LaAP-Net) to address this challenge. Our LaAP-Net not only generates the answer to the question but also predicts a bounding box as evidence of the generated answer. Moreover, a context-enriched OCR representation (COR) for multimodal fusion is proposed to facilitate the localization task. Our proposed LaAP-Net outperforms existing approaches on  three benchmark  datasets  for the text VQA task by a noticeable margin.

\end{abstract}

\section{Introduction}
\label{intro}

Visual Question Answering (VQA) has attracted much interest from the communities and witnessed tremendous progress. 
However, lacking the ability to generate answers based on texts in the image limits its applications. Recently, many new datasets \cite{biten2019icdar,textvqa} and new methods \cite{textvqa,hu2020iterative} are proposed to tackle this challenge and refer it as text VQA. 

The earliest method for text VQA is LoRRA\cite{textvqa}, which provides an optical character recognition (OCR) module for the VQA input and proposes a dynamic copy mechanism to select the answer from both fixed vocabulary and OCR words. The following work M4C\cite{hu2020iterative} inspired by LoRRA, uses rich representations of OCR as input and utilizes dynamic pointer network to deal with out-of-vocabulary answers, leading to state-of-the-art performance. However, M4C simply concatenates all modalities as transformer input and does not consider the high-level interaction among modalities of text VQA. Moreover, it is unable to provide evidence for the answer since the text is not localized in the image. 
Another recent work \cite{wang2020general} proposes a new dataset for evidence-based text VQA, which suggests Intersection over Union (IoU) based evaluation metric to measure the evidence.
Our work follows the spirit of evidence-based text VQA. More specifically, we generate the answer text bounding box during the answer prediction process as supplementary evidence for our answer. 
We propose a localization-aware answer prediction module (LaAP) that integrates the predicted bounding box with our semantic representation for the final answer. 
Besides, we propose a multimodal fusion module with context-enriched OCR representation, which uses a novel position-guided attention to integrate context object features into OCR representation. 

The contributions of this paper are summarized as follows: 
1) We propose a LaAP module, which predicts the OCR position and integrates it with the generated answer embedding for final answer prediction. 
2) We propose a context-enriched OCR representation (COR), which enhances the OCR modality and simplifies the multimodal input. 
3) We show that the predicted bounding box can provide evidence for analyzing network behavior in addition to improving the performance. 
4) Our proposed LaAP-Net outperforms state-of-the-art approaches  on  three  benchmark  text VQA datasets, TextVQA\cite{textvqa}, ST-VQA\cite{biten2019scene} and OCR-VQA\cite{ocrvqa}, by a noticeable margin.

\section{Related Works}

\subsection{Text Visual Question Answering}
Text VQA has attracted much attention from the communities.
The predominate method is LoRRA \cite{textvqa}, which takes image features, OCR features and questions to generate the answer.
LoRRA mimics the human answering process by providing the image-looking module, text-reading module and answer-reasoning module.  The generated answer could be selected from a fixed answer vocabulary or one of the OCR tokens by the copy module.  The copy module is further improved by M4C \cite{hu2020iterative} using dynamic pointer network. The M4C also proposes a transformer based network with 3 multi-modal input (question, image object features and OCR features). We share the same spirit as M4C but split the network into a clear encoder-decoder structure. 
We further propose a context-enriched OCR representation to extract OCR related image features.

\subsection{Evidence-based VQA and Multitask Learning}

Evidence-based VQA has been proposed in the recent work \cite{wang2020general}, which suggests to use intersection over union (IoU) to indicate the evidence. 
Many existing works
\cite{selvaraju2017grad,goyal2016towards,ref-mcan,ref-dynamic_fusion} compute the attention scores and build spatial maps on image to highlight regions, which the model focuses on. The spatial maps serve as an evidence and visual explanations of a VQA architecture.
Our method further extends this by designing a location predictor to generate a bounding box on the image to explain the answer generated. 
The bounding box explains that the correct answer generated is based on the analysis of underlying reasoning instead of exploiting the statistics of the dataset. As such, the bounding box becomes evidence of the VQA answer. 
To achieve the aforementioned target, we design a multitask learning process, which not only generates the answer based on the image and question but also provides the bounding box for the answer. 
The proposed method improves the interpretation of VQA results and leads to better performance.

\section{Localization-aware Answer Prediction Network}

\subsection{LaAP Network Architecture }

\begin{figure}[t]
	\centering 
	\includegraphics[width=0.95\textwidth]{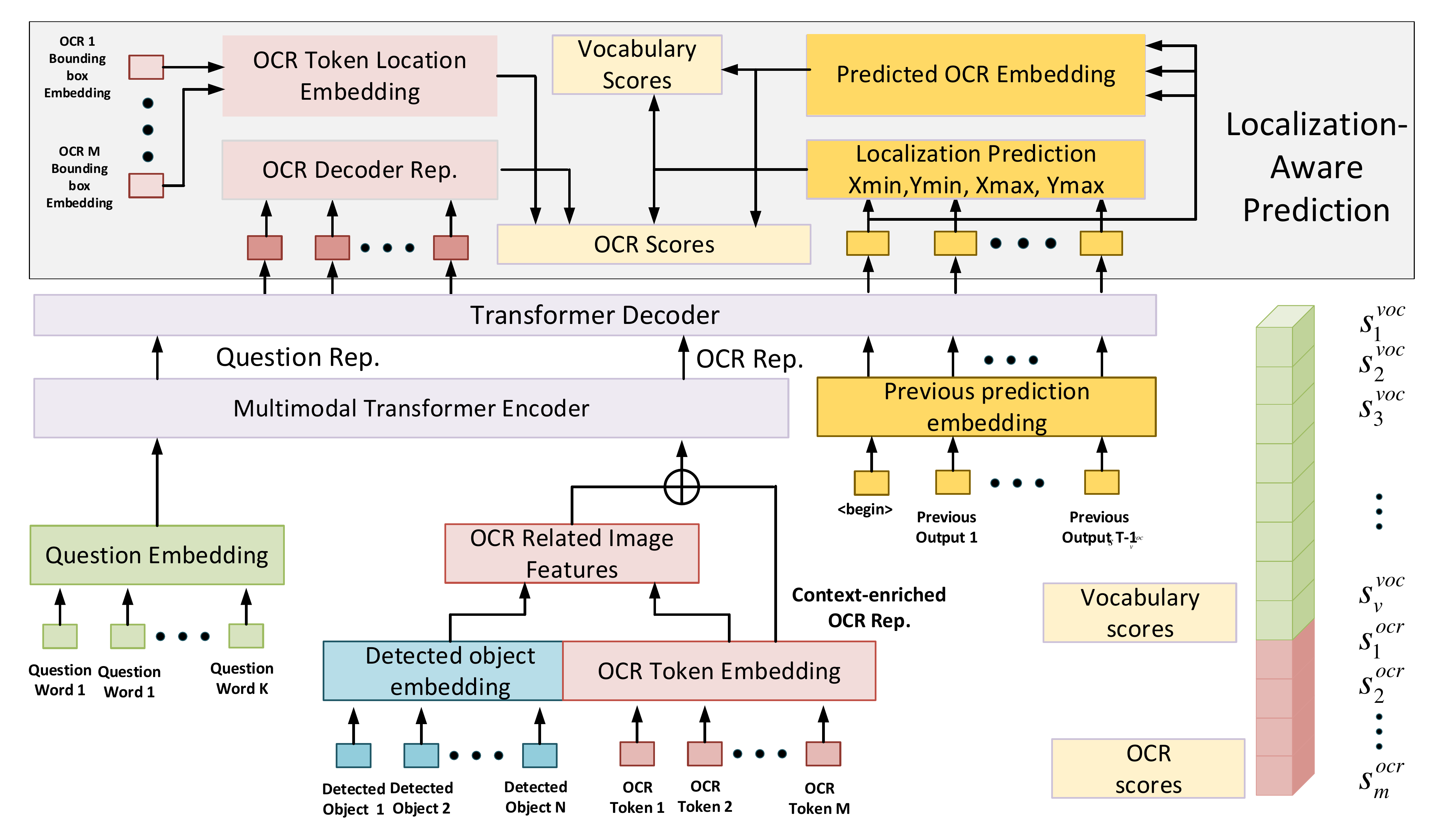}
	\caption{
		An overview of LaAP model.
		We perform context-enriched OCR representation to extrat object features. Then question words and enriched OCR tokens are input to the transformer encoder and the transformer decoder.  Based on the transformer decoder outputs, we first predict the answer localization, and then integrate this localization to the OCR embedding. Decoder output is also equipped with OCR position embedding. The OCR scores and vocabular scores are calculated accordingly to find the answer from an OCR token or a word from the fixed answer vocabulary.  
	}
	\label{fig:overall_arch}
\end{figure}

To better utilize the position information of image texts and enforce the network to better exploit visual features, we propose a localization-aware answer prediction network (LaAP-Net).
Our LaAP-Net is built based on the multimodal transformer encoder, transformer decoder and localization-aware prediction network as shown in Figure \ref{fig:overall_arch}. The transformer encoder takes the question embedding and OCR embedding as input.
Question embedding is generated by putting the question through a pretrained BERT-based model, whereas the OCR embedding is generated by our proposed context-enriched OCR representation module. 
As highlighted in dark yellow in Figure \ref{fig:overall_arch}, the decoding process starts with the \textit{<begin>} signal. For each decoded output, we first generate a bounding box.  This bounding box will then be embedded and added to the current answer decoder output, which is referred as localization-aware answer representation. Finally, it is fed to the vocabulary score module and OCR score module. The scores are concatenated and the element with the maximum score is selected as the final answer.
In the following section, we will present the three components of LaAP-Net: the context-enriched OCR representation, the localization-aware predictor and the transformer with simplified decoder.

\subsection{Context-enriched OCR Representation}

\begin{figure}[t]
	\centering 
	\includegraphics[width=0.9\textwidth]{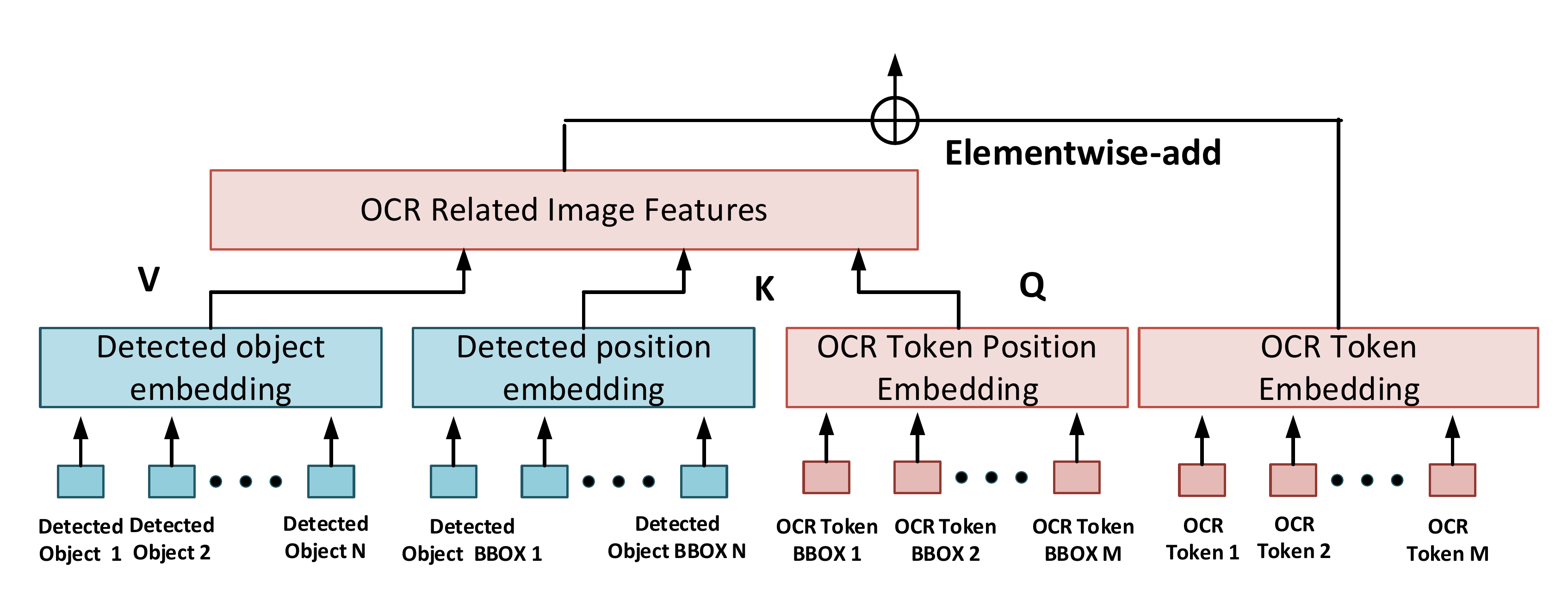}
	\caption{
	Context-enriched OCR representation to integrate the object features based on the spatial information (bounding box)
	}
	\label{fig:bbox_attn}
\end{figure}

Existing work \cite{hu2020iterative} builds a common embedding space for all modalities. However, this common embedding space has difficulty utilizing the image object features. We observe this by training the M4C \cite{hu2020iterative} network without the image object modality. The accuracy is almost unaffected.
To better exploit the image object modality, we propose the context-enriched OCR representation (COR) module (Shown in Figure \ref{fig:bbox_attn}). Ideally, the answer for text VQA should be found from OCR tokens, thus we integrate geometric context objects of an OCR token into its representation to improve the discriminative power. Take Figure \ref{fig:stvqa}(b) for example, the OCR representation for \textit{sixers} enriched with the features of the object \textit{red jersey} can be better attended by the question. The context objects are attended according to the proposed position-guided attention, where only spatial relationship between objects and OCR tokens are considered. 

Following M4C \cite{hu2020iterative}, we use features extracted from $N$ object detected by the Faster R-CNN \cite{ren2015faster}, denoted as $x_n^{obj} (\text{where } n=1,...,N)$. The corresponding bounding box coordinates are represented as $b^{obj}_n (\text{where } n=1,...,N)$. A combination of Faster R-CNN, Pyramidal Histogram of Characters (PHOC) \cite{almazan2014word} and FastText \cite{bojanowski2017enriching} embedding is adopted for $M$ OCR tokens in an image, denoted as $x_m^{ocr} (\text{where } m=1,...,M)$ with the bounding box denoted as $b^{ocr}_m (\text{where } m=1,...,M)$.  We embed the given question into a set of word embedding $x_k^{ques} (\text{where } k=1,...,K$ and $K$ is the number of words) through a pretrained BERT language model \cite{ref-bert}. All embeddings are then linearly projected to a $d$-dimensional space.

The detailed computation process for COR is described as follows. 
Firstly, the position-guided attention score vector $att_{m}$ between the $m$-th OCR token and the image objects is calculated as
\begin{equation}
    att_{m} = softmax((W^{Q}b^{ocr}_m)^T*[W^{K}b^{obj}_1, ..., W^{K}b^{obj}_N]), m=1,...,M
\end{equation}
where $W^{Q}$ and $W^{K}$ are query projection matrix and key projection matrix respectively.
Then the  $m$-th image attended OCR representation is calculated as weighted sum of the $N$ object feature vectors as
\begin{equation}
     x^{ocr|obj}_m = [x^{obj}_1, ..., x^{obj}_N]*att_{m}^T, m=1,...,M
\end{equation}
Note that we omit the multi-head attention mechanism \cite{ref-transformer} for simplicity. Finally, each OCR token is represented by aggregating OCR feature embedding, image attended OCR representation and position embedding as
\begin{equation}
   \hat{x}_{m}^{ocr} = x^{ocr}_m +  x^{ocr|obj}_m+ W^{ocr}b^{ocr}_m, m=1,...,M
\end{equation}
where $W^{ocr}$ is a matrix that linearly project the bounding box coordinate vector to $d$ dimension. With the proposed attention, the image object modality is merged into OCR.  We then feed {$ \hat{x}_{1}^{ocr}$,...,$\hat{x}_{M}^{ocr} $} and $x^{ques}_{1}, ..., x^{ques}_{K}$ into the transformer encoder as input. The strengthened OCR representation $ \hat{x}_{m}^{ocr} $ empowers the network to better learn the semantic correlation between OCR tokens and question. Meanwhile, it simplifies the multimodal feature input to improve the localization-aware answer prediction.

\subsection{Localization-aware Answer Prediction}
\label{lap}
\subsubsection{Localization-aware Predictor }

To exploit the positional information of image features and texts, we design a localization-aware predictor to perform the bounding box prediction. The bounding box is embedded and added to the decoder output to generate the localization-aware answer representation.
More specifically, given the answer embedding $y^{dec}$ output from the decoder, we calculate the localization-aware answer representation $z^{ans}$ by fusing $y^{dec}$ with the gated bounding box projection as
\begin{equation}
\label{eq:LAR}
z^{ans} = y^{dec} + g^{loc} \circ (W^{loc}*b^{pred}+bias^{loc})
\end{equation}
where $W^{loc}$ and $bias^{loc}$ are weights of a linear layer to project the location bounding box to the same dimension as $y^{dec}$ and $\circ$ represents element-wise multiplication. $g^{loc}$ is the localization gate. Note that our network update the gate weight automatically through training, so that it implicitly reveals the statistical importance of the localization information. 
Similarly, we calculate the high-level localization-aware representation $z^{ocr}_{m} (\text{where } m=1,...M)$ of each OCR token as
\begin{equation}
z^{ocr}_{m} = y^{ocr}_{m} + g^{loc} \circ (W^{loc}*b^{ocr}_{m}+bias^{loc}),\  m=1,...,M
\end{equation}
where $y^{ocr}_{m}, $ denotes the m-th OCR encoding from the last encoder layer and $b^{ocr}_{m}$ is the corresponding bounding box coordinates. $b^{ocr}_{m}$ goes through the same linear projection layer and localization gate as $b^{pred}$ so that they are projected to the same high-dimensional space. \\
Then similar to \cite{hu2020iterative}, we obtain the similarity score $s^{ocr}_m$ between each OCR representation and the answer representation as 
\begin{equation}
\label{eq:ocrScore}
s^{ocr}_{m} = (W^{ans}*z^{ans}+bias^{ans})(W^{ocr}*z^{ocr}_{m}+bias^{ocr}), \  m=1,...,M
\end{equation}
where $W^{ans}$, $bias^{ans}$, $W^{ocr}$ and $bias^{ocr}$ are parameters of linear projection layers. 
The localization-aware answer representation $z^{ans}$ is also fed into a classifier to output $V$ scores $s^{voc}_v \ (v=1,...,V)$, where $V$ is the vocabulary size. The final prediction is selected as the element with the maximum score as
\begin{equation}
\max\: [s^{ocr}_m; s^{voc}_v]
\end{equation}

Note that the predicted bounding box is not explicitly used in generating the answer. However, localization prediction is a vision task so it can enforce the network to exploit visual features. As a result, it serves as a good complement to the classical vocabulary classification task,  which mainly focuses on linguistic semantics. The localization-aware predictor strengthens the learned answer embedding to attend to the correct OCR token, which in turn facilitates the classifier to correctly find the word. 
Moreover, this localization information improves the performance of position-related questions as shown in Figure \ref{fig:texvqaIm}(a) and \ref{fig:texvqaIm}(c), which will be further discussed in Section \ref{evalTextvqa}

\subsubsection{Loss Design to Incorporate the Evidence Scores }
We use the IoU scores as the evidence for the answer generated. Therefore, we propose a multitask loss, which facilitates the answer embedding to learn both the semantics and localization information provided by the OCR tokens. The proposed multi-task loss consists of three individual loss functions: localization loss $L_l$, semantic loss $L_s$ and the fusion loss $L_f$.

The answer embedding output from the decoder is fed into a multilayer perceptron (MLP) to directly predict the bounding box location $b^{pred}$ of the answer OCR token. Inspired by \cite{trans_detr}, the localization loss $L_{l}$ is defined as:
\begin{equation}
    L_{l} = (1-IoU(b^{pred}, b^{gt}) + \mathds{L}_1(b^{pred}, b^{gt}))*\mathds{I}
\end{equation}
where $b^{gt}$ denotes the ground truth bounding box, which is obtained by matching the OCR token text to the ground truth answer text. $IoU$ and $\mathds{L}_1$ calculate the intersection over union and L1 norm respectively between the prediction and ground-truth bounding box. $\mathds{I} = 1$ if the answer word matches one of the recognized OCR text and 0 otherwise.

To accurately answer a question, OCR localization and semantic information are both critical. Thus, we propose a fusion loss $L_f$ to couple the localization prediction and semantic representation of the answer. The two aspects of information complement each other in the process of decision making.
Formally, given the target scores $t^{ocr}_{m}\in \{0,1\} (m=1,...M)$, we formulate our fusion loss $L_f$ using cross entropy as
\begin{equation}
\label{eq.fusionLoss}
    L_f = \sum_{i=1}^{M}  loss_{cross\_entropy}(s^{ocr}_{m},  t^{ocr}_{m})
\end{equation}

In order to exploit the linguistic meaning of the answer embedding, we collect a fixed vocabulary of frequently used words. We feed the localization-aware answer representation $z^{ans}$ into a linear classifier to classify answer embedding of each decoding step to one of the vocabulary. Our semantic loss $L_s$ is computed as the cross entropy between the classification score vector and the one hot encoding from the ground truth word. The overall multi-task loss of the network is calculated as
$L = L_f + \lambda_{l}L_l + \lambda_{s}L_s$,
where $\lambda_{l}$ and $ \lambda_{s}$ are regulation coefficients that determine the importance of localization loss and semantic loss. The value of $\lambda_{l}$ and $ \lambda_{s}$ are experimentally selected.

\begin{figure}[t]
	\centering 
	\includegraphics[width=0.9\textwidth]{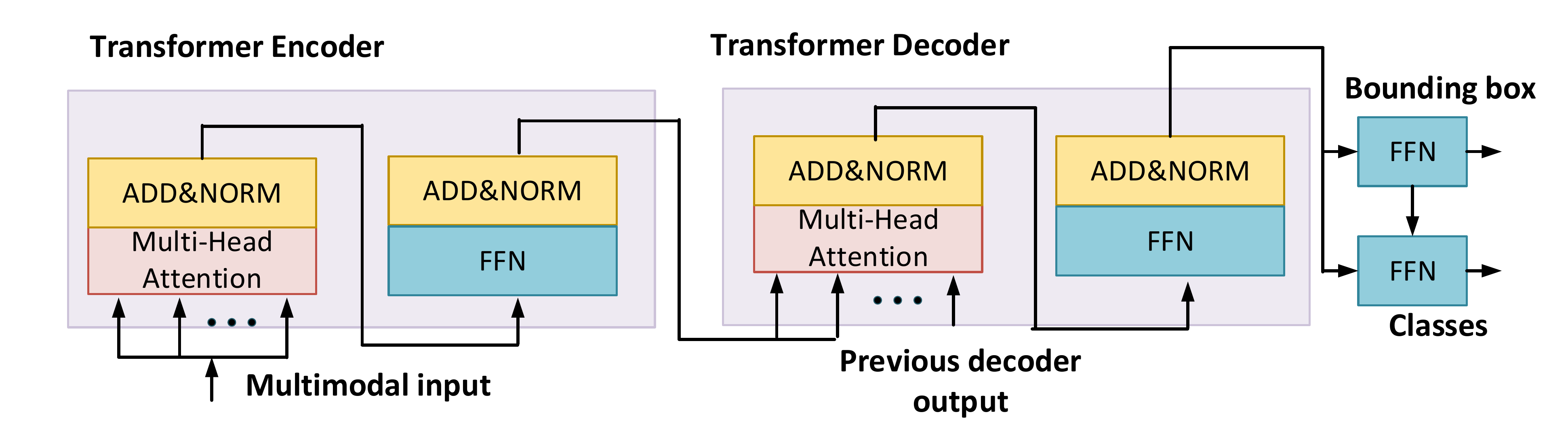}
	\caption{An overview of the transformer with simplified decoder (TSD). TSD output is used to generate the bounding box, which is then used for answer prediction
	}
	\label{fig:enc_dec}
\end{figure}

\subsection{Transformer with Simplified Decoder}
Existing works \cite{hu2020iterative,gao2020structured} use BERT alike transformer architecture, which allows each decoder layer to attend to the same depth encoder layer. 
However, a deeper encoder layer extracts a more broad view of the input than a shallow layer \cite{clark2019does}. As such, we adopt the standard transformer encoder-decoder structure as shown in Figure \ref{fig:enc_dec}. Here, we use the transformer with simplified decoder (TSD) by removing the decoder self-attention to save the computational cost. We experimentally find that only using the encoder-decoder attention can maintain the same performance.
The multimodal inputs are encoded by $L$ stacked standard transformer encoder layers. The embedding of the last encoder layer is fed into each of the $L$ decoder layers.
The answer word is generated in an auto-regressive manner, i.e. for each decoding step, we take the predicted answer embedding from the previous step as the decoder input and obtain the answer embedding as the decoder output.
The decoding process is performed by the proposed localization-aware prediction module as shown in Figure \ref{fig:overall_arch} and discussed in Section \ref{lap}.

\section{Experiments}
\label{sec:length}
We evaluate our LaAP-Net on the three challenging benchmark datasets: TextVQA \cite{textvqa}, ST-VQA \cite{biten2019icdar} and OCR-VQA \cite{ocrvqa}. 
We show that the proposed LaAP-VQA network outperforms state-of-the-art works on these datasets. We further perform the ablation study to investigate the proposed context-enriched OCR representation (COR) and the localization-aware answer prediction (LaAP) on TextVQA dataset.  

\input{table/textvqa_abalation}

\subsection{Implementation Details}
For a fair comparison with the state-of-the-art methods, we follow the same  multimodal input
as M4C \cite{hu2020iterative}.  More specifically, we use a pretrained BERT \cite{ref-bert} model for question encoding, the Rosetta-en OCR system \cite{borisyuk2018rosetta} for OCR representation and a Faster-RCNN \cite{ren2015faster} based image feature extraction.  The OCR tokens are represented by a concatenation of the appearance features from Faster R-CNN , FastText embeddings \cite{bojanowski2017enriching} , PHOC feature \cite{almazan2014word} and bounding box (bbox) embedding. 
We set the common dimensionality $d=768$ and the number of transformer layers $L=4$.
More details of training configuration are summarized in the supplementary material.


\begin{figure}[thb]
	\centering
	\setlength{\abovecaptionskip}{0.1cm}
	\includegraphics[width=0.99\textwidth]{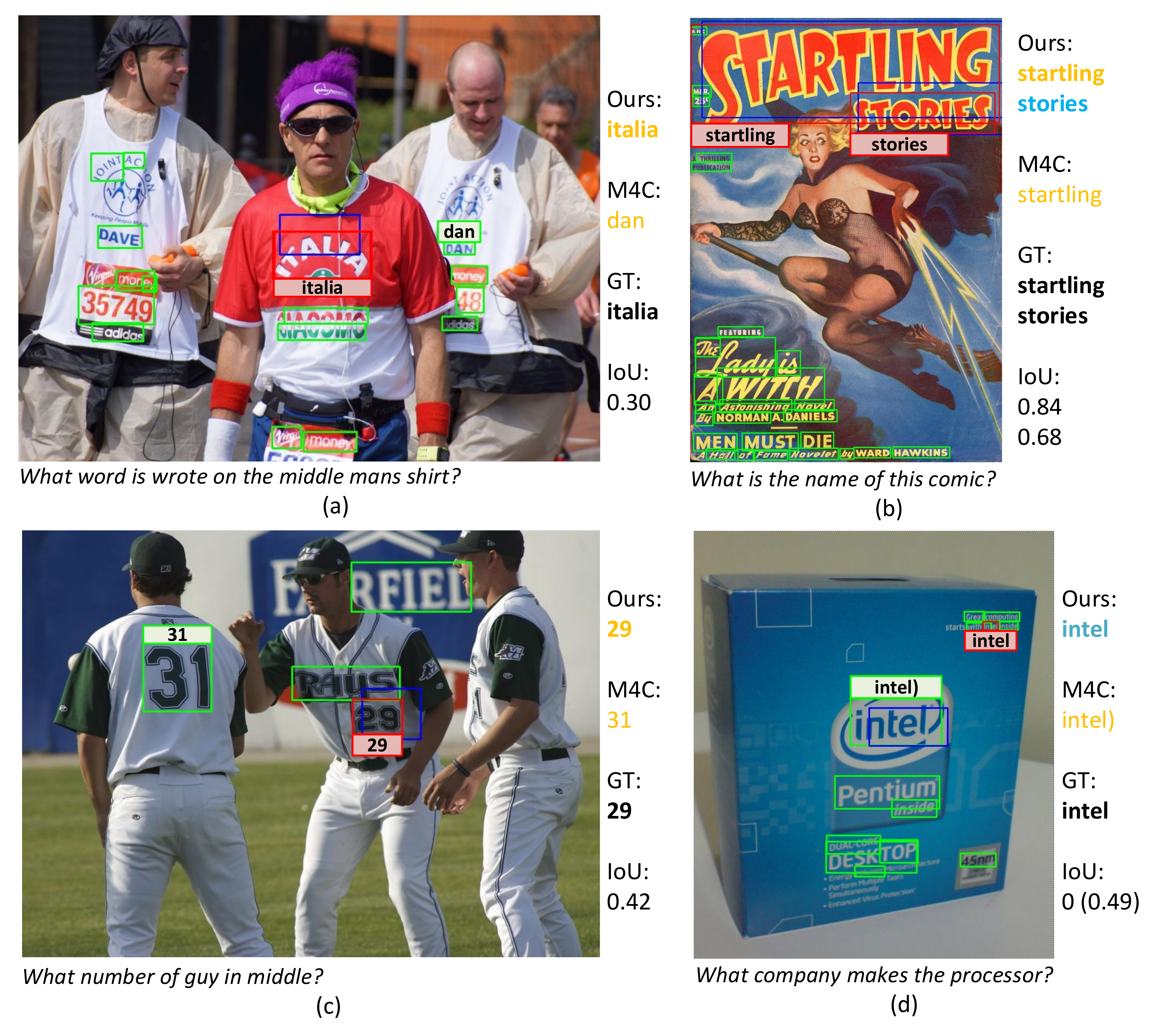}
\\
	\caption{Qualitative examples from TextVQA dataset. We display predicted answers (Yellow for word generated from OCR and blue for vocabulary) of LaAP-Net and M4C with ground-truth (GT). Our predicted bounding box (blue box) is also depicted in the images to compare to the GT box (red box). The IoU as an evidence score is also shown in each image. For Figure (d), IoU 0 (0.49) indicates OCR recognition error, 0.49 is the IoU with the GT bounding box.  }
	\label{fig:texvqaIm}
\end{figure}

\subsection{Evaluation on TextVQA Dataset}
\label{evalTextvqa}

The TextVQA \cite{textvqa} dataset contains 28408 images with 34602 training, 5000 validation and 5734 testing question-answer pairs.
We compare our result on TextVQA to the newest SOTA method SMA \cite{gao2020structured} and other existing works like LoRRA \cite{textvqa}, MSFT VTI \cite{msf_vti}, and M4C \cite{hu2020iterative}. The proposed LaAP-Net achieves a 40.68\% validation accuracy and a 40.54\% testing accuracy, which improves the SOTA by 1.10\% (absolute) and 0.25\% (absolute). Note that we only compare with SMA results using the same set of features to show the advantage of the network structure itself. We also train our network with additional data from ST-VQA dataset following M4C and boost the test accuracy by 0.95\% (absolute). \\
\textbf{Ablation Study on Network Components}.   Context-enriched OCR representation (COR) and Localization-aware predictor (LaP) are the two key features of our network. We investigate the importance of both components by progressively adding them on our transformer with simplified decoder (TSD) backbone. First, we remove COR and LaP from our network and feed image object feature directly into the encoder as in M4C. The answer prediction part is also strictly following M4C. This configuration is denoted as TSD in Table.\ref{tab:abalation}. Then we add COR on TSD, which is denoted as TSD+COR. The third ablation is adding only LaP to TSD (TSD+LaP). Each component demonstrates a contribution to performance improvement as shown in Table \ref{tab:abalation}. To further prove the effectiveness of COR and LaP, we add them on our baseline network M4C. COR and LaP individually lead to an accuracy improvement of 0.38 and 0.95 respectively. COR and LaP together boost the accuracy by 1.33.  Note that our network without COR, i.e. TSD+LaP surfers from performance retrogress. The rationale behind is that flat multimodal feature used in place of COR contains both objects and OCR tokens. Object's position embedding introduces much noise for the localization task. COR absorbs context object features in OCR representation and improves its discriminating power. Meanwhile, the encoder multimodal input is simplified, which leads to noise reduction. In summary, LaP and COR are not two independent modules simply added together. They enhance each other and improve our network as a system.\\
\textbf{Ablation Study on Source of Answer}. We restrict the answer generation source to study the effect of our method on word semantic learning and OCR selection. As shown in Table.\ref{tab:layer}, our model significantly improves the accuracy when we only predict the answer from vocabulary. It implies that our localization prediction module enhances the network's capacity for learning the semantics of OCR tokens, which coincides with our qualitative analysis.\\
\textbf{Evidence-based Qualitative Analysis on TextVQA Dataset.}
One challenge for the existing VQA system is that the correct answer generated is hard to tell whether the answer is based on the analysis of underlying reasoning or through exploiting the statistics of the dataset. As such, Intersection-over-Union (IoU) \cite{wang2020general} is recommended to measure the evidence for the answer generated.
The IoU result of our bounding box is shown in Figure \ref{fig:texvqaIm}.  For example, in Figure  \ref{fig:texvqaIm}(b), two IoU results (0.84, 0.68) explain the reason for the answer "startling stories".
Higher IoU indicates better evidence. Furthermore, these IoU scores show the answer is generated by exploiting the image features instead of exploiting the statistics of the data set, i.e. a coincidental correlation in the data.

Furthermore, we observe that most of the text VQA errors come from inaccurate OCR result.  e.g. in Figure \ref{fig:texvqaIm}(d), the OCR token "intel)" is recognized wrongly, which results in the false answer of M4C. Due to the localization prediction, our method generates the correct answer even in such case (\ref{fig:texvqaIm}(d)). Since localization tends to use visual features of OCR tokens rather than their text embedding, it can better determine the attended OCR token in spite of the text recognition result.
With the predicted OCR bounding box, the answer generation problem is converted to a conditioned classification process $P(\text{text}|\text{predicted box})$ to recognize the text from the vocabulary. More examples supporting our analysis can be found in Figure \ref{fig:texvqaIm}.

 Our localization predictor also shows the capability of understanding position and direction as shown in Figure \ref{fig:texvqaIm}(a, c). Our network learns to understand position in training because the ground-truth position is provided straight to guide the localization prediction, while in previous works, positional information is put through several layers of encoder and decoder without explicit guidance. 

\input{table/textvqa_cmp}

\begin{figure}[t]

    \centering
    \setlength{\abovecaptionskip}{0.2cm}
\subfloat[]{	    \includegraphics[height=0.38\textwidth]{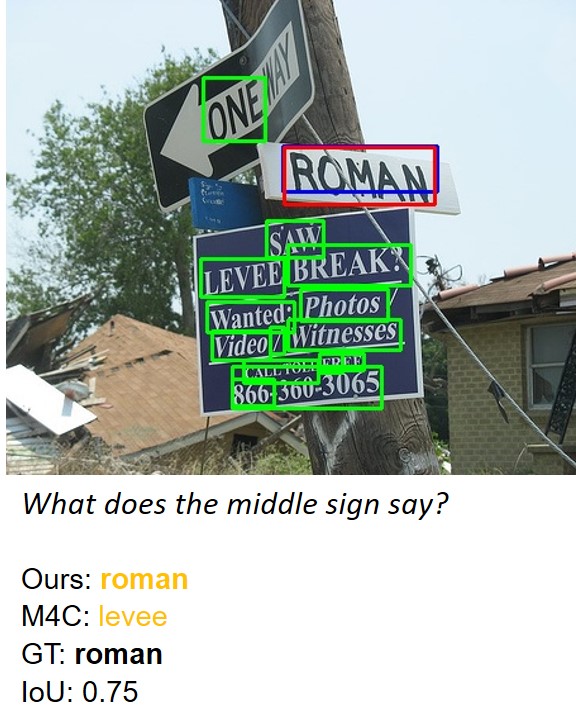}}
\subfloat[]{    \includegraphics[height=0.38\textwidth]{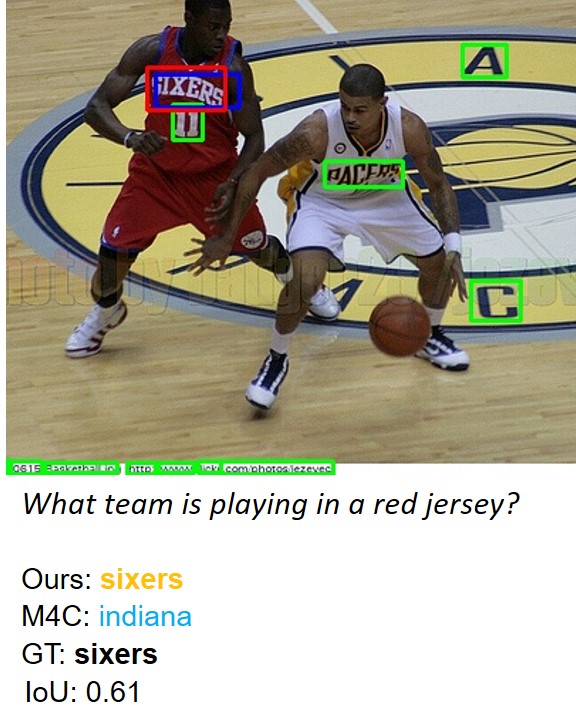}}
\subfloat[]{    \includegraphics[height=0.38\textwidth]{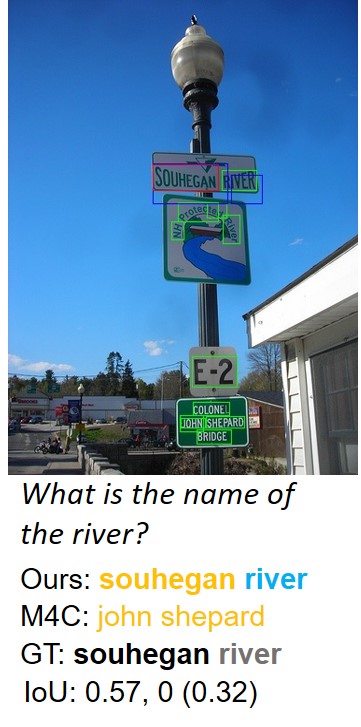}}
\subfloat[]{    \includegraphics[height=0.38\textwidth]{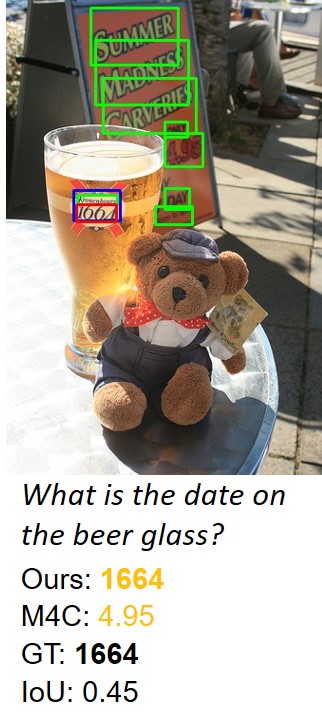}}
    \caption{Qualitative examples from ST-VQA dataset. We display predicted answers (Yellow for word generated from OCR and blue for vocabulary) of LaAP-Net and M4C with ground-truth (GT) (GT word not contained in any OCR  is printed in gray). Our predicted bounding box is also depicted in the images (blue box) to compare to the GT box (red box). The IoU as an evidence score is also shown in each image. For Figure (c), IoU 0 (0.32) indicates OCR recognition error, 0.32 is the IoU with the GT bounding box. }
   \label{fig:stvqa}
\end{figure}

\subsection{Evaluation on ST-VQA Dataset.}
We evaluate the proposed model on the open vocabulary task of ST-VQA \cite{biten2019scene}, which contains 18921 training-validation images and 2971 test images. Following previous works \cite{hu2020iterative,gao2020structured}, we split the images into training and validation set with size of 17028 and 1893 respectively.

We report both accuracy and ANLS score (default metric of ST-VQA) in Table \ref{tab:stvqa}. Our LaAP-Net surpasses the SOTA method by a large margin on both metrics. Note that SMA improves its baseline method M4C by only 0.004 in testing ANLS score while we boost the result by 0.019. \\
\textbf{Evidence-based Qualitative Analysis on ST-VQA}
Figure \ref{fig:stvqa} shows IoU scores, our predicted bounding box and answer. In those examples, our proposed localization-aware answer predictor not only generates correct answer, but also predicts exact bounding box(drawn in blue) of the corresponding OCR token. Similar conclusion can be drawn from the result as discussed for TextVQA dataset. In Figure \ref{fig:stvqa}(a), our network correctly attends to the middle sign designated by the question, where our reference method M4C fails. In Figure \ref{fig:stvqa}(c), our network manages to predict the word 'river' even though it is not recognized by the OCR system.  More qualitative examples can be found in the supplementary material.

\input{table/stvqa_ocrvqa}

\subsection{Evaluation on OCR-VQA Dataset}
Unlike TextVQA and ST-VQA that contain "in the wild" images,  OCR-VQA dataset consists of 207572 images only of book covers. Thus, the image object modality is less important in OCR-VQA.
Moreover, since questions are about the title or author of a book, it is relatively difficult to determine the location. Even so, our model still achieves the state-of-the-art result, 64.1\% accuracy as shown in Table \ref{tab:ocrvqa}. 

\subsection{Failure Analysis}
Two failure cases are shown in Figure \ref{fig:false}. As discussed in Section \ref{evalTextvqa}, our model is sensitive to positional instruction in a question. However, in Figure \ref{fig:false}(a), the question asks about relative position, which our network does not gain knowledge on. In Figure \ref{fig:false}(b), the position "right" is indicated by an arrow, but our network locates the road sign on the right of the image. In this case, question answering requires reasoning in addition to text reading function, which we will investigate in our future work.

%
%
%
%
%


\section{Conclusion}
This paper proposes a localization-aware answer prediction network (LaAP-Net) for text VQA.  Our LaAP-Net not only generates the answer to the question, but also provides a bounding box as an evidence of the answer generated. Moreover, a context-enriched OCR (COR) representation is proposed to integrate object related features. The proposed LaAP-Net outperforms existing  approaches  on  three  benchmark  datasets  
for the text VQA task by a noticeable margin with new state-of-the-art performance: TextVQA 41.41\% , ST-VQA 0.485 (ANLS) and OCR-VQA 64.1\%. 

\bibliographystyle{coling}
\bibliography{coling2020}



\section*{Supplementary material}

# Appendix

## A Implementation and Hyper-parameter Setting

We use the LoRRA open sourced training environment developed by PyTorch. The question encoder is performed by the first 3 layer of the pre-trained BERT-base model. We also set a maximum decoder step 12 in the answer prediction process. Answer length 12 covers almost all the answers for TextVQA, ST-VQA and OCR-VQA datasets.

We build a fixed vocabulary by using the top 5000 frequent words from the answers of the TextVQA dataset. TextVQA has 10 answers for each question, and the accuracy is measured by the soft-voting for the ten answers. The ST-VQA official metric is average normalized levenshtein similarity (ANLS), which is defined as scores $1 - d_L(a_{pred}, a_{gt})/max(|a_{pred}|, |a_{gt}|)$, (where $a_{pred}$ and $a_{gt}$ are prediction and ground-truth answers respectively, and $d_L$ is the edit distance) averaged over all questions.

The training batch size is set to 64 with 32000 iterations for both TextVQA and ST-VQA dataset. For the OCR-VQA dataset, the batch size is 128. The best validation accuracy model is used for the test dataset. The entire training takes around 7 hours for TextVQA, 11 hours for ST-VQA and 11 hours for OCR-VQA on 2 Nvida 2080ti GPUs. We summarize the hyper-parameter table for the reproducibility purpose.

| Input Feature Parameters | Value | Optimizaiton Parameters | Value |
|---|---|---|---|
| max question word number K | 20 | optimizer | Adam |
| max detected object number N | 100 | warm-up learning rate factor | 0.2 |
| image object feature dimension | 2048 | max grad L2 -norm for clipping | 0.25 |
| max input OCR token number M | 50 | warm-up iteration | 1000 |
| OCR Fasttext dimension | 300 | max iteration TextVQA | 32000 |
| OCR PHOC dimension | 604 | max iteration ST-VQA, OCR-VQA | 48000 |
| OCR Faster R-RCNN dimension | 2048 | learning rate steps, TextVQA, ST-VQA | 14000, 19000 |
| input embedding dimension d | 768 | learning rate steps, OCR-VQA | 28000, 38000 |
| input dropout rate | 0.1 | base learning rate | 1e-4 |
| batch size, TextVQA, ST-VQA | 64 | learning rate decay | 0.1 |
| batch size OCR-VQA | 128 | max decoding steps T | 12 |
| Text BERT layer | 3 | LaAP-net Encoder/Decoder Layer | 4 |

## B Additional Qualitative Examples

In this section, we show additional examples of text VQA answer and predicted bounding box from both TextVQA and ST-VQA datasets to support our claim in the paper.

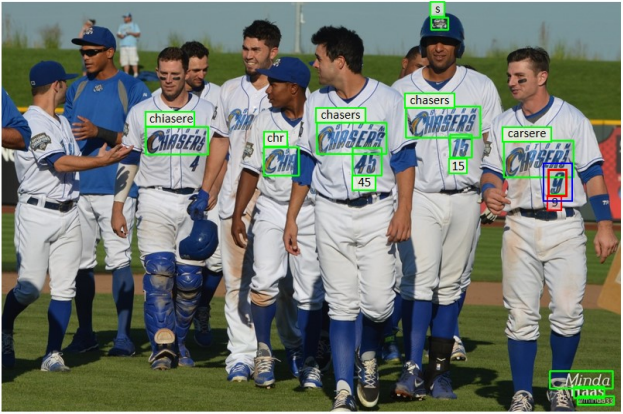
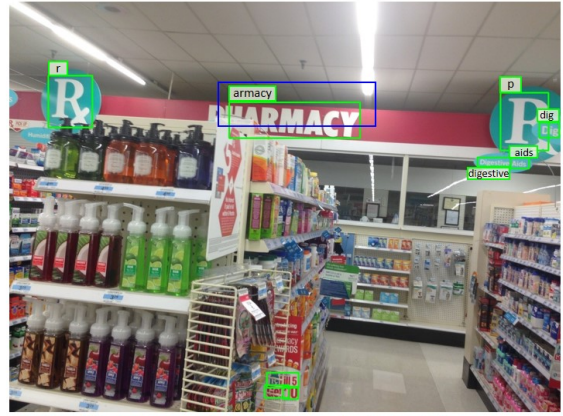

*What is the number of the player furthest right?*
Ours: **9**    GT: **9**

(a)

*What department is being shown at the store?*
Ours: **pharmacy**    GT: **pharmacy**

(b)

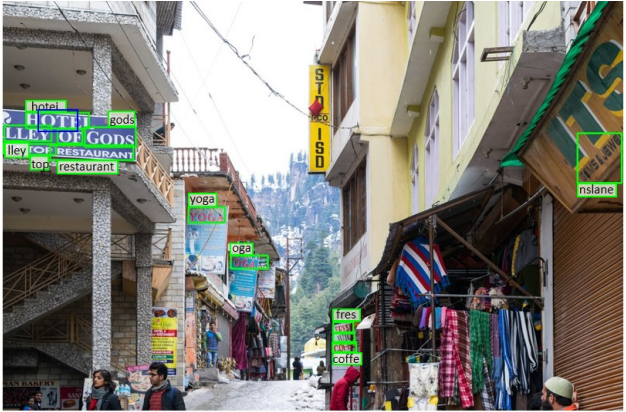
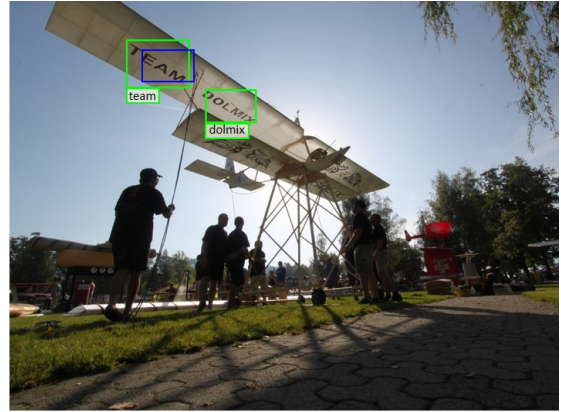

*Is there a hotel in this area?*
Ours: **yes**    GT: **yes**

(c)

*Are they part of team dolmix?*
Ours: **yes**    GT: **yes**

(d)

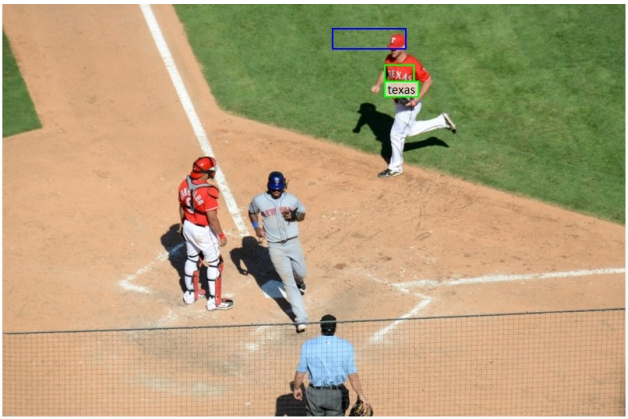
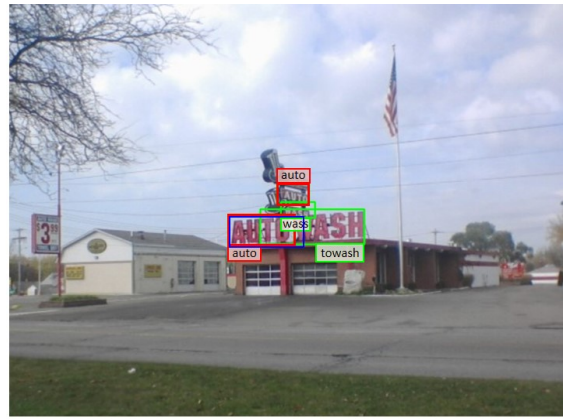

*What letter is on the red hat?*
Ours: **t**    GT: **t**

(e)

*What kind of wash is available?*
Ours: **auto**    GT: **auto**

(f)

**Figure 1:** Qualitative examples from TextVQA dataset. We display predicted answers (Yellow for word generated from OCR and blue for vocabulary) of our LaAP-Net and the ground-truth (GT). Our predicted bounding box (blue box) is also depicted in the images to compare to the GT box (red box). Note that some images do not contain GT bounding box while some images contain more than one GT bounding box

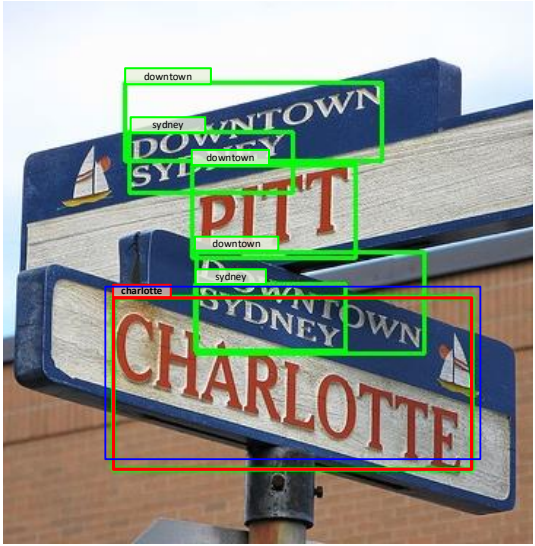 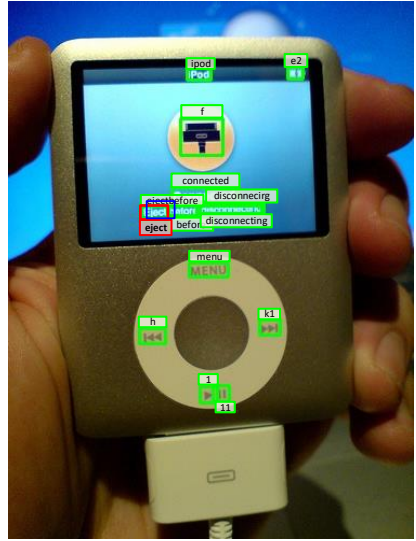

*What is the bottom street name?*
Ours: **charlotte**    GT: **charlotte**
(a)

*What do you have to do before disconnecting?*
Ours: **eject**    GT: **eject**
(b)

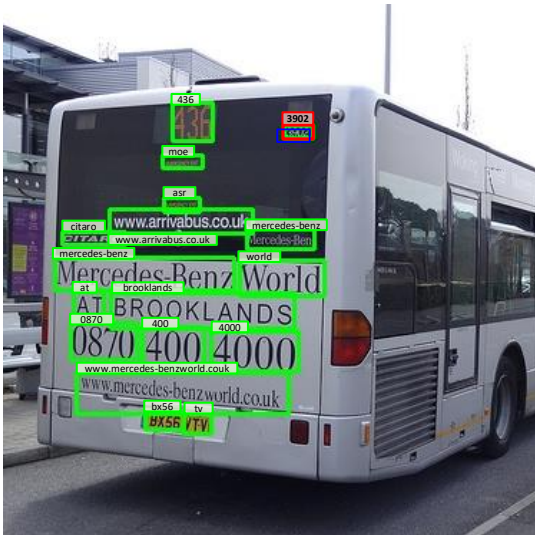 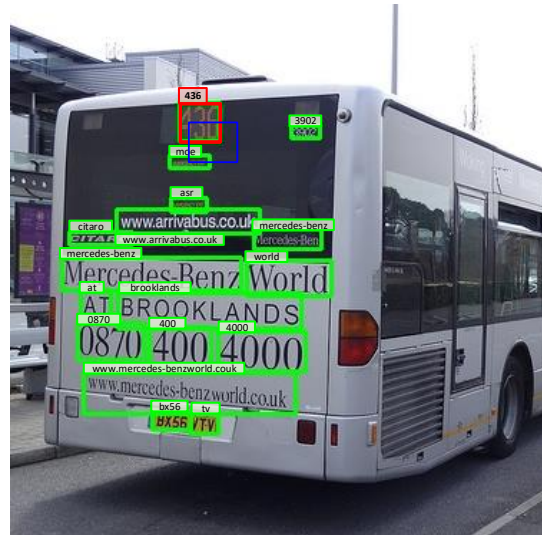

*What is the white number on the top right of the back of the bus?*
Ours: **3092**    GT: **3092**
(c)

*What is the digital number in red at the top of the bus?*
Ours: **436**    GT: **436**
(d)

**Figure 2:** Qualitative examples from ST-VQA dataset. We display predicted answers (Yellow for word generated from OCR and blue for vocabulary) of our LaAP-Net and the ground-truth (GT). Our predicted bounding box (blue box) is also depicted in the images to compare to the GT box (red box).



\end{document}

%% file: table/textvqa_abalation.tex
\begin{table}[b]

\parbox[t]{.45\linewidth}{
\centering
\scalebox{0.95}{
\begin{tabular}{|c|c|c|}
\hline
\textbf{Methods}          & \textbf{Val Acc.} & \textbf{Test Acc.} \\ \hline
LoRRA 			 & 26.56       & 27.63               \\ \hline
M4c               & 39.40       & 39.01             \\ \hline
M4C+COR            &     39.78   & ----              \\ \hline
M4C+COR+LaP &            40.73        &      40.41        \\ \hline \hline
TSD             &     39.86              & ----              \\ \hline
TSD+LaP        &     39.37              & ----          \\ \hline
TSD+COR       &     40.49             & ----              \\ \hline
TSD+COR+LaP(ours)                                & 40.68            & 40.54             \\ \hline
\end{tabular}
}
\caption{\label{tab:abalation} Ablation study on context-enriched OCR representation and localization-aware answer prediction for M4C model and our proposed model}
}
\hfill
\parbox[t]{.45\linewidth}{
\centering
\scalebox{0.95}{
\begin{tabular}{|c|c|}
\hline
\textbf{Methods}                                                          & \textbf{Val Acc}  \\ \hline
M4C        & 39.40                  \\ \hline
w/o Vocab       & 31.76                  \\ \hline
w/o OCR copy       & 14.94                  \\ \hline \hline 
LaAP-Net (ours)        & 40.68                  \\ \hline
w/o Vocab.                                                              &       31.37              \\ \hline
w/o OCR Copy                                                              &       24.71               \\ \hline
w/o OCR embedding Copy                                                        &     34.51               \\ \hline
w/o OCR bbox Copy                                                              &       40.49        \\ \hline
\end{tabular}
}
\caption{\label{tab:layer}
	Ablation study by removing its fixed answer vocabulary (w/o Vocab.) or  OCR copying (w/o OCR Copy) on the TextVQA dataset.}
}
\end{table}

%% file: table/textvqa_cmp.tex
\begin{table}[t]
\centering
	\parbox[t]{.45\linewidth}{
\scalebox{0.9}{
\begin{tabular}{|c|c|c|c|c|}
\hline
Method                                            & \begin{tabular}[c]{@{}c@{}}Val. \\ Acc. \end{tabular} & \begin{tabular}[c]{@{}c@{}}Test \\ Acc. \end{tabular} \\ \hline
LoRRA \cite{textvqa}                                   &   26.56                   & 27.63            \\ \hline
MSFT VTI \cite{msf_vti}                                & 32,92                     & 32.46                                                  \\ \hline
M4C \cite{hu2020iterative}                               & 39.4                                   & 39.1                                                   \\ \hline
SMA  \cite{gao2020structured}                            & 39.58                                                  & 40.29                                                  \\ \hline
LaAP-Net                                                & \textbf{40.68  }                                                & \textbf{40.54  }                                                \\ \hline \hline
M4C + STVQA \cite{hu2020iterative}                      & 40.55                                  & 40.46                                                   \\ \hline
 LaAP-Net + STVQA                                         &              \textbf{41.02}     &   \textbf{41.41 }         \\ \hline
\end{tabular}
}

\caption{\label{tab:textqa} 
On the TextVQA dataset, our model outperforms LoRRA and M4C by 13.11\%  and 1.44\% (absolute) respectively. Our final model trained on TextVQA and STVQA dataset advances the state-of-the-art performance to 41.41\% on TextVQA test split dataset. }
}
	\hfill
	\parbox[t]{.45\linewidth}{
		\scalebox{0.9}{
			\begin{tabular}{|c|c|c|c|}
				\hline
Method            & \begin{tabular}[c]{@{}c@{}}Acc.  \\ Val\end{tabular} & \begin{tabular}[c]{@{}c@{}}ANLS \\  Val.\end{tabular}  & \begin{tabular}[c]{@{}c@{}}ANLS \\  Test \end{tabular} \\ \hline				

				\begin{tabular}[c]{@{}c@{}} VTA \\ \cite{biten2019icdar}  \end{tabular} 
				& -----           & -----       & 0.282        \\ \hline
				\begin{tabular}[c]{@{}c@{}} M4C \\ \cite{hu2020iterative}  \end{tabular}        
				& 38.05           & 0.472       & 0.462        \\ \hline
				\begin{tabular}[c]{@{}c@{}} SMA \\ \cite{gao2020structured} \end{tabular} 		                         & -----           & -----       & 0.466        \\ \hline
				LaAP-Net                    & 39.74           & 0.4974       & \textbf{0.485}       \\ \hline
			\end{tabular}
		}
		\caption{\label{tab:stvqa} On the STVQA datset, our LaAP-Net model achieves +0.02 (absolute) ANLS over the most recent work SMA and approximately +0.2 (absolute) boost over the challenge winner, VTA \cite{biten2019icdar}. }
	}

\end{table}

%% file: table/stvqa_ocrvqa.tex
\begin{figure}[!t]
\setlength{\abovecaptionskip}{0.2cm}
  \begin{minipage}{\textwidth}
	\begin{minipage}[b]{0.5\textwidth}
		\centering
		\subfloat[]{
         	\includegraphics[height=0.65\textwidth]{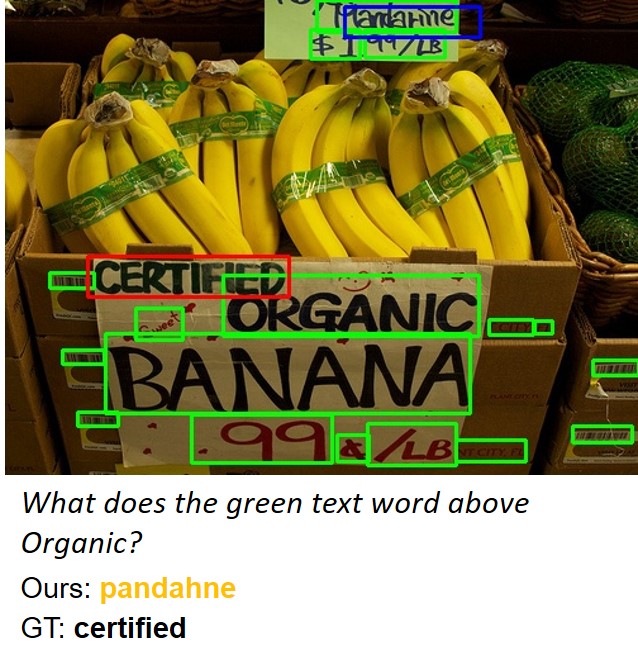}}
         \subfloat[]{
	    	\includegraphics[height=0.65\textwidth]{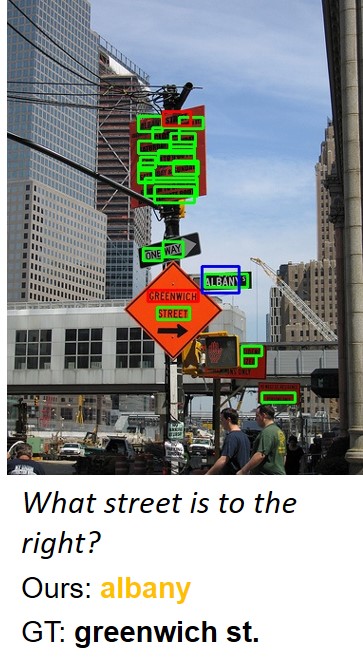}}
		\captionof{figure}{Failure examples of our LaAP-Net}
		\label{fig:false}
	\end{minipage}
	\hfill
	\begin{minipage}[b]{0.4\textwidth}
		\scalebox{0.9}{
		\begin{tabular}{|c|c|c|}
		\hline
		Method                                                 & Val Acc. & Test Acc. \\ \hline
		\begin{tabular}[c]{@{}c@{}}BLOCK\\ \cite{ocrvqa} \end{tabular}     &   ----       & 42.0      \\ \hline
		\begin{tabular}[c]{@{}c@{}}CNN\\ \cite{ocrvqa} \end{tabular}       &   ----       & 14.3      \\ \hline
		\begin{tabular}[c]{@{}c@{}}Combine+W2V\\ \cite{ocrvqa} \end{tabular} &  ----        & 48.3      \\ \hline
		\begin{tabular}[c]{@{}c@{}}M4C\\
		\cite{hu2020iterative}  \end{tabular}       & 63.5     & 63.9      \\ \hline
		LaAP-VQA                                               & \textbf{63.8 }    & \textbf{64.1 }     \\ \hline 
	\end{tabular}
}
\\

		\captionof{table}{\label{tab:ocrvqa}  On the OCR-VQA dataset, our LaAP-VQA model achieves the state-of-the-art result 64.1\% accuracy comparing to M4C \cite{hu2020iterative}.}
	\end{minipage}
\end{minipage}
\end{figure}